\pdfoutput=1
\documentclass[11pt]{article}

\usepackage{ACL2023}

\usepackage{times}
\usepackage{latexsym}

\usepackage[T1]{fontenc}
\usepackage[utf8]{inputenc}

\usepackage{microtype}

\usepackage{inconsolata}

\usepackage{graphicx}       % import graphical images
\usepackage{multirow}       % merge rows in tables
\usepackage{subcaption}     % create sub-tables and sub-figures
\usepackage{bold-extra}     % enable \texttt{\textbf{}}
\usepackage{bm}             % enable bold font in math mode
\usepackage[hang,flushmargin]{footmisc}  % minimize footnote indentation
\usepackage{stfloats}  % for bottom-aligned table
\usepackage{amsfonts,amsmath,amssymb}
\usepackage{pifont}
\usepackage{array,booktabs,makecell,tabularx}
\usepackage{lipsum}
\usepackage{enumitem}
\usepackage{xcolor}

\newcommand{\cmark}{\ding{51}}
\newcommand{\xmark}{\ding{55}}

\title{Towards Open-World Product Attribute Mining:\\A Lightly-Supervised Approach}

\author{Liyan Xu$^{1,2}$\thanks{\;\;Primary work done as an applied scientist intern at Amazon, before joining WeChat AI.} \quad Chenwei Zhang$^{3}$ \quad Xian Li$^{3}$ \quad Jingbo Shang$^{4}$ \quad Jinho D. Choi$^{2}$ \\
$^{1}$Pattern Recognition Center, WeChat AI \qquad $^{2}$Emory University\\
$^{3}$Amazon.com \qquad
$^{4}$University of California, San Diego\\
\normalsize{\texttt{liyan.xu@emory.edu} \quad \{{\texttt{cwzhang,xianlee}\}\texttt{@amazon.com}}}
}

\begin{document}
\maketitle

% \footnotetext{\textcolor{red}{\textbf{Unpublished draft. Not for distribution.}}}

\begin{abstract}
% \lipsum[1-1]

% \jingbo{general comment: sentences are too long to parse}
We present a new task setting for attribute mining on e-commerce products, serving as a practical solution to extract open-world attributes without extensive human intervention.
Our supervision comes from a high-quality seed attribute set bootstrapped from existing resources, and we aim to expand the attribute vocabulary of existing seed types, and also to discover any new attribute types automatically.
A new dataset is created to support our setting, and our approach Amacer is proposed specifically to tackle the limited supervision.
Especially, given that no direct supervision is available for those unseen new attributes, our novel formulation exploits self-supervised heuristic and unsupervised latent attributes, which attains implicit semantic signals as additional supervision by leveraging product context.
Experiments suggest that our approach surpasses various baselines by 12 F1, expanding attributes of existing types significantly by up to 12 times, and discovering values from 39\% new types.
Our data and code can be found at \url{https://github.com/lxucs/woam}.
\end{abstract}
\section{Introduction}
\label{sec:intro}

% closed-world vs. zero-shot vs. open-world
% distant-supervision vs. weak-human-supervision

Attribute mining (or product attribute extraction) is to extract values of various attribute types (e.g. \textit{colors}, \textit{flavors}) from e-commerce product description, which is a foundational piece for product understanding in online shopping services, enabling better search and recommendation experience.

Within this task regime, different settings have been studied. Most pioneer works deem it as a closed-world setting, where models are trained to identify a fixed set of pre-defined attribute types \cite{ghani-attr-mining,putthividhya-hu-2011-bootstrapped,opentag}, similar to the standard named entity recognition (NER). Recent works start to step up towards the open-world aspect that supports extraction of new attribute types unseen in training. Particularly, several works have focused on the zero-shot perspective \cite{su-opentag,mave}, enabling extraction of a new attribute type during inference if given a name or description of this new type, which is a more realistic setting to this task, as new types of products and attributes are constantly emerging in the real world.

\begin{figure}[t]
\centering
\includegraphics[width=\columnwidth]{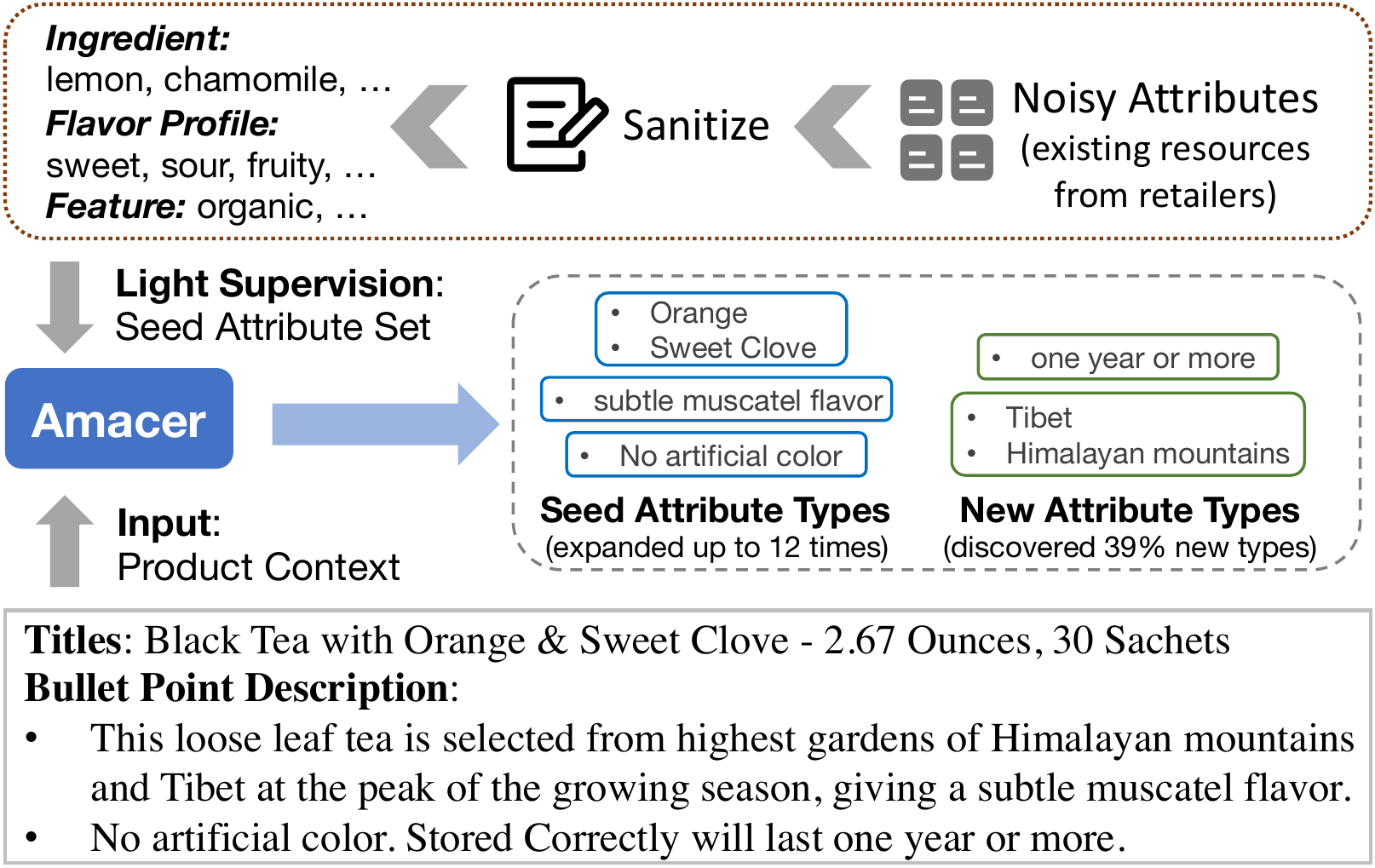}
\caption{Illustration of our task setting on one product: given light supervision from seed attributes, our approach Amacer aims to expand attribute vocabulary of seed types, and to also discover values of any new types (\textit{Shelf Life}, \textit{Origin}) not covered by seeds. The outputs on all products are thus attribute clusters with diverse values. Evaluation is based on clustering metrics, as new clusters are not named beforehand.}
% \chenwei{make sure the color scheme is consistent with fig 2, and ideally the same set of example values being used}}
% \jingbo{For the ``Human Curation'', is it more like a simple sanity check as a filtering process? If so, I would recommend ``Human Filtering'' and explain how easy it is. ``Curation'' sounds complicated}
% \jingbo{In the output part, it is not clear if we are distinguishing/naming the attribute types?}
\label{fig:task}
\vspace{-2ex}
\end{figure}

In this work, we formulate the attribute mining task one step further towards the ultimate open-world setting: given product-related description, the objective is to identify as many new values of existing attribute types, as well as any new types that could be considered as reasonable attributes but not covered in training. As such, our setting automatically discovers new attributes, unlike the zero-shot setting that requires explicit specification of new types of interest. In addition, we also aim the model to work under limited supervision, by introducing only a relatively small seed attribute set in training, thereby remaining practical when only a few values are known for a certain attribute, also for the fact that it is untenable to keep up high-coverage human annotations of ever-changing attributes, especially in e-commerce domain.

Figure~\ref{fig:task} illustrates our overall task setting, where the model expands the attribute vocabulary of existing types, and discovers any new attributes, yielding numerous attribute clusters. A new dataset dubbed \textsc{WoaM} (\textbf{W}eakly-supervised \textbf{O}pen-world \textbf{A}ttribute \textbf{M}ining) is created to accommodate our setting, as described in Section~\ref{sec:data}.
Targeting towards realistic open-world setting, our dataset covers full product horizons including titles and detailed description, where the latter provides rich context and is shown to contain more unseen attribute types than titles by 66\% (Table~\ref{tab:dataset-brief}).
% The most similar work to our setting is OA-Mine \cite{oamine}, which also adopts the open-world and weakly-supervised paradigm, but under a limited scope that only extracts from product titles. We increase the scope to include full description for each product, which provides much more diverse context and is shown to contain more unseen attribute types than titles by 66\% (Table~\ref{tab:dataset-brief}).
Moreover, distinguished from previous datasets that either require substantial annotation efforts \cite{opentag} or noisy distant-supervised data \cite{su-opentag,mave,oamine}, our training supervision comes from a high-quality seed attribute set constructed hybridly, combining data-driven and light human curation.
Overall, our setting achieves good trade-offs with reasonable human interventions, under a practical scope with decent coverage on attributes.

We then propose our approach for this setting, dubbed \textbf{Amacer} (\textbf{A}ttribute \textbf{m}ining with \textbf{a}daptive \textbf{c}lustering and w\textbf{e}ak \textbf{r}egularization).
To overcome the challenge of limited supervision, we first introduce our approach to generate diverse spans of candidate attribute values from corpus (\textsection\ref{sec:attribute-generation});
% \jingbo{Instead of using \textsection, I'd prefer to just say Section}
% from Liyan: this is to save some space
then focus on representation learning by utilizing explicit supervision from seed attributes (\textsection\ref{sec:seed-attr}), followed by the last step that performs grouping on candidate spans using refined, attribute-aware embeddings (\textsection\ref{sec:grouping}). New formulations to mine more implicit semantic signals from product context are also proposed for new attribute discovery (\textsection\ref{sec:new-attr}).
% It consists of three stages: extracting candidate spans of plausible values; attribute-aware representation learning on span embedding space; grouping candidate spans into attribute clusters.

Experiments on \textsc{WoaM} suggest that our approach outperforms various baselines by up to 12.5 F1. Furthermore, our novel formulation to leverage self-supervised and unsupervised semantic signals is shown effective to both existing and new attributes, especially boosting new attribute discovery by a good margin of 6.4 F1.
% \chenwei{looks impressive! wondering on which metric and what does such improvement implies}. 
% Furthermore, we propose to leverage self-supervised and unsupervised methods as weak regularization that reveals additional semantic signals, which has not been explored previously, and brings another 2.2 F1 improvement.
Despite the limited amount of seed values, our model is able to expand the seed attribute vocabulary by up to 12 times (Table~\ref{tab:result-example}), and to discover values from 39\% unseen attribute types on our test set.
% our setting can produce practical predictions on seed attributes (Table~\ref{tab:result-example}), while the open-world discovery remains a future research problem that needs to be further refined (Table~\ref{tab:analysis-cmp},\ref{tab:result-example-new})\chenwei{<-i am more comfortable putting this here when we discuss the major areas for improvements explicitly in the paper (and argue we are doing a pilot study on this challenging real-world problem), like what typical survey papers did}.
Overall, our contributions can be summarized as follows:
% \jingbo{shall we split the dataset into the second contribution? so three contributions in total: new problem, new data, new method}
\begin{itemize}[noitemsep,nolistsep,leftmargin=*]
    \item We address a new setting in attribute mining as a practical paradigm to extract open-world attributes under light human intervention.
    \item A new dataset is created, covering 66 attribute types with 42\% unseen types from the seed set.
    \item A new approach is proposed to support our unique task setting, especially exploiting self-supervised and unsupervised semantic signals, which has not been explored by previous works.
\end{itemize}

\section{Data}
\label{sec:data}

Our dataset \textsc{WoaM} consists of three parts, including: 1) text corpus; 2) seed attribute set for training; 3) human-annotated test set for evaluation. Full statistics of our dataset are provided in Table~\ref{tab:dataset}, and more details are provided in Appendix~\ref{apdx:dataset}.

% Our new dataset has two main differences compared to the dataset released by \citet{oamine}: (1) we constructed the seed attribute set of significantly higher quality (described in Section~\ref{subsec:dataset-seed}) and higher coverage (shown in Table~\ref{tab:dataset}); (2) we make use of product bullet point description in addition to titles, which can provide richer information on product attributes. In Table~\ref{tab:title-bp}, we further provide more characteristics of bullet points compared to titles. In particular:

% \begin{itemize}[leftmargin=*]
%     \item Bullet point description is shown to be 32.3\% longer than titles on-average, while there are less supervision signals from seed (fewer seed value occurrences), indicating that stronger inference capability is required for the model to extract attributes from bullet points, motivating our approach to leverage self-supervised and unsupervised task signals from the text corpus (described in Section~\ref{subsec:attribute-reg}).
%     \item Bullet point description has more human-identified attributes, especially with a higher ratio of new attribute types, which demonstrates that attribute discovery from bullet points can be more practical and diverse to mine richer attribute expressions from text corpus.
% \end{itemize}

\paragraph{Corpus}
Four common e-commerce product categories are included in our corpus: \textit{Tea, Vitamin, Sofa, Phone Case}.
For each category, we sampled 9,000+ products publicly listed on Amazon.com with full description available in English. Each product record can be represented as a tuple: (identifier, category, title, bullet points).
% \chenwei{do we really needed all those notations (including the notations for each bullet)? saw that we didn't refer to all of them later} $p_i = (d_i, c_i, t_i, B_i)$, where $d_i$ is the unique i\textbf{d}entifier, $c_i$ is the corresponding product \textbf{c}ategory, $t_i$ is the product \textbf{t}itle, and $B_i = \{b_{i_1}, .., b_{i_k}\}$ covers $k$ corresponding \textbf{b}ullet point description.

\paragraph{Seed Set}
For each category, the seed set consists of a few applicable attribute types (avg. 16.5 types per category) and their values (avg. 22 values per type).
We adopt a hybrid approach for the construction: existing resources are first utilized to bootstrap the seed set, and human curation is performed upon to overcome the noisy issue existed in previous datasets (example shown in Table~\ref{tab:oamine-example}).
Specifically, two steps are applied as below:

% It is built with three objectives in mind: (1) serving as weak supervision for training; (2) being scalable, thereby practical to add more categories in the future; (3) of high quality, to overcome the noisy issue existed in previous distant-supervised datasets. Therefore, we adopt a hybrid approach: we first bootstrap the seed set by utilizing existing resources \jingbo{it will be nice to name a few other resources like Walmart, Target, or even ChatGPT?}, then perform human curation upon it to balance between the speed and quality.

% Specifically, we make use of the product profiles \jingbo{add a figure to illustrate what are the product profiles. The same figure can be used to demonstrate the three drawbacks mentioned in the later sentences. Stats in Appendix A are nice but an example is strongly recommended.} collected from Amazon.com, which can be viewed as a knowledge source containing records of certain attributes mostly filled out by product sellers. The raw profiles are not ideal to use as training supervision directly due to three drawbacks: noisy attributes, data sparsity, coarse granularity (more details in Appendix~\ref{apdx:dataset}).
% To address those issues, we constructed the seed set by two phases \jingbo{phases, not phrases. It is a popular typo and see if there are more places like this}:

\textit{\textbf{Automatic Sanitizing}}:
we collect the raw product profiles that contain certain attributes provided by Amazon retailers, and perform frequency-based heuristics to heavily sanitize noisy attributes.
First, long-tail attribute types that have fewer than 10 values are removed. Second, for each product category, if a unique value appears under multiple attributes types, we restrict it to only belong to its most common type. Lastly, for each attribute type, we only keep at most 100 values based on the top frequency, so to discard the tail values that we are less confident on. The resulting seed set thereby has a relatively small size but of higher quality after above three steps.
% Each resulting attribute keeps at most 100 popular values. \jingbo{it sounds like a hard frequency threshold? Shall we share this number?}

\textit{\textbf{Human Curation}}:
as the attribute set after sanitizing is relatively small, human curators can go through the entire set rather quickly and consolidate the final seed set ($<40$ min per product category).
Concretely, remaining noisy values are spotted and removed from their attribute types. Furthermore, granularity is adjusted such that ambiguous or coarse attribute types are split into multiple newly defined fine-grained types; similar attribute types are also merged into one type.

After we obtain the final seed set, we perform string match to obtain their occurrences in corpus, ready to be used for training. A development set is separately created that consists of sanitized profile attributes solely for hyperparameter tuning.
Overall, our training supervision is built practically that balances between scalability and quality. 

\paragraph{Test Set}
For each category, we collect additional products not covered in the raw corpus as the test set. Two in-house annotators 
% \jingbo{what are the agreement ratios between these two annotators?} 
are asked to annotate all spans that appear as reasonable attribute values of either an existing type from the seed set, or a brand-new type that fits the context.
As with previous works, we do not allow overlapping spans: more complete spans are preferred over shorter and incomplete spans; each span is assigned a single attribute type that best describes its property.

% From Table~\ref{tab:dataset}, we can see that around 87\% attribute values in the test set are not covered by the seed set, with 42\% attribute types being completely new types. \chenwei{as a reader i wish we cover more of this in this section -- sacrificing other details seems not a dealbreaker (e.g.granularity issues; less than 40 min per category ->} More insights of our dataset, including its unique challenges and comparisons, are further documented in Appendix~\ref{apdx:dataset}.

Table~\ref{tab:dataset-brief} briefly specifies unique characteristics of our dataset. It is clear that most gold values are new values unseen from the seed set. Especially, bullet points have a higher ratio of new attribute types/values than titles, while those values are harder to extract due to longer text, sparser values, and more complex language structures.
For comparison, our setting poses greater challenges than the most related previous dataset from a recent work OA-Mine \cite{oamine}, which is under a much limited scope that consists of only titles with sparser and noisier seed attributes (detailed comparison is provided in Appendix~\ref{apdx:prev}).

\begin{table}[t!]
\centering
\resizebox{\columnwidth}{!}{
\begin{tabular}{l|cccc}
\toprule
& Type (New) & Value (New) & Tok & Gold \\
\midrule
\tt TT & 46 (28\%) & 864 (70\%) &  20.1 & 5.7 (28.5\%) \\
\tt BP & 65 (43\%) & 2787 (89\%) &  26.6 & 3.6 (13.8\%) \\
\bottomrule
\end{tabular}}
\caption{Characteristics of our dataset by titles (\texttt{TT}) and bullet points (\texttt{BP}) on the test set (full stats in Table~\ref{tab:dataset}): total number of unique attribute types/values, with the ratio of new types/values in parentheses; averaged number of tokens and gold values per title/bullet sequence, with the density of gold values per token in parentheses.}
\label{tab:dataset-brief}
\vspace{-1ex}
\end{table}

\begin{figure*}[t]
\centering
\includegraphics[width=\textwidth]{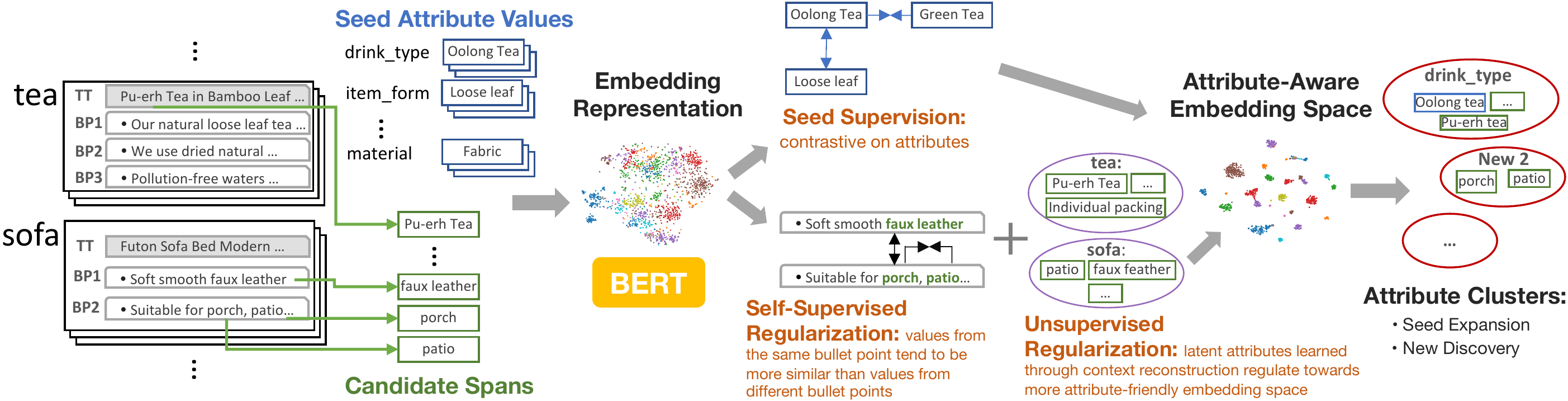}
\caption{Illustration of our proposed approach Amacer. It generates candidate spans from product description (\textsection\ref{sec:attribute-generation}), and performs representation learning on embedding space, by utilizing: explicit supervision from seed attributes (\textsection\ref{sec:seed-attr}); implicit semantic signals by self-supervised heuristic and unsupervised latent attributes (\textsection\ref{sec:new-attr}). Final attribute clusters can be obtained by grouping candidates through adaptive expansion and DBSCAN (\textsection\ref{sec:grouping}).}
\label{fig:model}
\vspace{-2ex}
\end{figure*}

Our proposed approach for this dataset is presented in the following Section~\ref{sec:attribute-generation}-\ref{sec:new-attr}. Specifically, Section~\ref{sec:attribute-generation}-\ref{sec:grouping} introduce the overall pipeline depicted in Figure~\ref{fig:model} that utilizes explicit signals from seed attributes, and Section~\ref{sec:new-attr} introduces our novel formulation to exploit implicit signals beyond the limited seed attributes.

% The pipeline largely follows the OA-Mine framework presented in \cite{oamine} that achieves state-of-the-art performance in a similar task setting. \chenwei{the argument can be less intrinsic to appreciate and lacks technical motivation. consider tying each ``upgrade'' we had with certain challenges so that people may appreciate them more ->} Nevertheless, we adopt completely different method for each stage (Section~\ref{subsec:attribute-generation}-\ref{subsec:attribute-grouping}) that is shown advantageous upon the previous framework, and we further propose to leverage self-supervised and unsupervised regularization to learn implicit task patterns (Section~\ref{subsec:attribute-reg}), in addition to the supervision signals from the seed set.

\section{Candidate Span Generation}
\label{sec:attribute-generation}

% Our proposed approach \textbf{Amacer} can be viewed as a three-stage pipeline: 1) Candidate Span Generation, which extracts certain spans from text corpus as candidate values; 2) Representation Learning on span embedding space, such that values of similar attributes have a closer embedding representation; 3) Candidate Span Grouping on candidate values, producing attribute clusters based on the obtained attribute-aware embedding similarity.

% As Section~\ref{subsec:attribute-generation}-\ref{subsec:attribute-grouping} introduces our entire pipeline, we further propose to enhance representation learning by leveraging self-supervised and unsupervised regularization to learn additional semantic signals, later described in Section~\ref{subsec:attribute-reg}.

The first stage of our approach is to generate spans from product description that could be qualified as attribute values, producing a set of non-overlapping candidate spans, serving as a foundational step for this attribute extraction task.

% This process aims to capture as many gold attribute values as possible, seeking to retain both common values and rare long-tail values. Thus, it emphasizes on recall over precision and allows certain noise, as later grouping stage will further discard noisy spans.
With weak supervision in mind, this step should not simply rely on signals from the seed set; otherwise, it would become hard to generalize and lose diverse attribute expressions during inference. Therefore, directly employing a supervised model can be suboptimal. It is also tempting to use off-the-shelf phrase extraction tools such as AutoPhrase \cite{autophrase}, however, the domain shift on e-commerce description of varied categories can severely affect recall, as observed by \citet{oamine}. 
The close work OA-Mine regards this stage as an unsupervised sentence segmentation task on product titles through language model probing \cite{wu-etal-2020-perturbed}, regarding each segment as a candidate span. Nonetheless, two shortcomings still remain. First, unlike titles, segmentation may not be suitable for bullet points, as most segments from bullet points would be noisy spans, demonstrated by the lower value density (13.8\%) in Table~\ref{tab:dataset-brief}.
% \chenwei{<-better when we mention which metrics we are relying on to draw this conclusion?}
% unlike titles that are concise and have little redundant information.
Second, being completely unsupervised, there is no task-specific adjustment in this process, suffering inadequate candidate quality.

In this work, we instead resort to a basic yet effective strategy that overcomes above issues, by using \textbf{syntax-oriented patterns}: we collect valid Part-of-Speech (POS) patterns for attribute values, and simply obtain all spans in the corpus that fit into those patterns as candidate spans, followed by rudimentary stopword filtering and overlapping span removal (prioritizing longer spans), yielding a smaller but higher-quality candidate set than that from sentence segmentation.

Valid POS patterns are acquired in a data-driven fashion without human intervention:
we leverage the product profiles again, and obtain all POS sequences of their attribute values. These raw sequences are further compacted by removing consecutive duplicate POS tags, such that \textit{healthy clean water} ([\texttt{ADJ}, \texttt{ADJ}, \texttt{NOUN}] → [\texttt{ADJ}, \texttt{NOUN}]) will share the same POS pattern as \textit{clean water} ([\texttt{ADJ}, \texttt{NOUN}]).
The resulting set of collected POS patterns serves to identify spans as well-formed or ill-formed phrases.

Examples of our POS patterns are shown in Table~\ref{tab:pos}. They regulate spans based on their syntactic features, without sole reliance on semantic supervision from the limited seed set, hence being able to capture diverse attribute expressions of vast variety. Overall, they serve as the quality guardrail for candidate spans, while reaping additional advantages: 1) easy to perform manual domain-specific adjustment; 2) scalable towards other product categories, as being data-driven; 3) efficient to run in practice.

\begin{table}[ht!]
\centering
\resizebox{\columnwidth}{!}{
\begin{tabular}{lll}
\toprule
\it healthy clean water & \small [\texttt{ADJ}, \texttt{NOUN}] & \cmark \\
\it sweet and spicy taste & \small [\texttt{ADJ}, \texttt{CCONJ}, \texttt{ADJ}, \texttt{NOUN}] & \cmark \\
\it promotes healthy liver function & \small [\texttt{VERB}, \texttt{ADJ}, \texttt{NOUN}] & \cmark \\
\it are available during & \small [\texttt{VERB}, \texttt{ADJ}, \texttt{ADP}] & \xmark \\
\it freshness so every cup & \small [\texttt{NOUN}, \texttt{ADV}, \texttt{DET}, \texttt{NOUN}] & \xmark \\
\bottomrule
\end{tabular}}
\caption{Examples of POS patterns to recognize well-formed (\cmark) or ill-formed (\xmark) phrases.}
\label{tab:pos}
\vspace{-1ex}
\end{table}

As we depend on external tools to identify POS, this process is not without noises. Nonetheless, we find the empirical performance to be quite robust qualitatively. Moreover, it can be augmented with other techniques to mitigate noise in scenarios tailored to specific applications.

% \section{Supervision from Seed Attributes}
% \section{Explicit Seed Supervisions for Vocabulary Expansion}
\section{Explicit Signals for Seed Expansion}
\label{sec:seed-attr}

With both seed attribute values and candidate spans in-place, 
our next objective is to perform representation learning that refines the geometry of embedding space, such that values of similar attributes should have a closer embedding representation, and vice versa, as the key property to leverage in later grouping stage. 
In this section, we introduce the utilization of available seed attributes as \textbf{explicit supervision}, primarily targeting the vocabulary expansion of existing attribute types.

For each seed value or candidate span, we can have an initial representation on the embedding space via encoding through pretrained language models such as BERT \cite{devlin-etal-2019-bert}. Concretely, we feed each text sequence (either a title or bullet point) to BERT, 
% where each sequence is also prepended by its corresponding product category followed by \texttt{[SEP]}, so that the encoding is explicitly conditioned on the category, useful to distinguish category-specific attributes.
and obtain the contextualized representation of each span by averaging its token embedding, without introducing extra encoding parameters.

% \subsubsection{Supervised Contrastive Learning}
% \label{subsec:attr-supervised}
% \chenwei{this subsubsection is the only one under its parent}

\paragraph{Supervised Contrastive Learning}
Contrastive learning is a natural fit to consume task signals from the seed set: for an anchor seed value $v_a$, a positive seed $v_p$ from the same attribute, and a negative seed $v_n$ from a different attribute, contrastive learning enforces $(v_a, v_p)$ to be more similar than $(v_a, v_n)$ on the embedding space.
OA-Mine adopts a triplet loss \cite{triplet-loss} for the supervised contrastive learning, as well as another regression loss \cite{sentence-bert} that directly pushes the similarity of positive/negative pairs, requiring careful sampling and tuning. In our work, we simplify this supervised process by only using an in-batch negative contrastive loss \cite{supervised-contrastive}. Let $I^s$ be all seed value indices, $P^s(i)$ be the indices of positive seeds that belong to the same attribute as seed $i$, $N^s(i) = I^s \setminus P^s(i)$ be the corresponding negative seeds. $g_i$ is the L2-normalized embedding of seed $i$ from the last layer of BERT encoding. The loss can then be denoted as:
\begin{align}
% \label{eq:attr-sup}
    \mathcal{L}^{su} = \sum_{i \in I^s} \frac{-1}{|P^s(i)|} \sum_{p \in P^s(i)} \log \frac{e^{(g_i \cdot g_p / \tau)}}{\sum_{j \in N^s(i)} e^{(g_i \cdot g_j / \tau)}} \nonumber
\end{align}
$\tau$ is the temperature hyperparameter.
As all embeddings are L2-normalized, $g_i \cdot g_j$ is effectively the cosine similarity as a distance measurement of two span representation.
$\mathcal{L}^{su}$ pushes seed values of the same attribute to have a similar representation, while pulling away seed values from different attribute types on the embedding space.

% \subsection{Candidate Span Grouping}
% \label{subsec:attribute-grouping}

\section{Candidate Span Grouping}
\label{sec:grouping}

After representation learning, a grouping stage upon candidate spans is followed. Each resulting cluster represents an attribute type, with each span inside being its attribute value.
Unlike most related works that employ off-the-shelf clustering algorithms such as HAC, K-Means or DBSCAN \cite{elsahar-open-re,zhao-etal-2021-relation,oamine},
% OA-Mine also ensembles results from DBSCAN with another attribute classification layer to increase recall on existing attributes.
% In our approach, we propose a two-step clustering strategy that directly utilizes the seed set, which is shown higher performance without introducing extra parameters:
% (1) grouping existing and new attributes separately to enable a direct view of results on each part\chenwei{<- does that also help achieve better performance -- what is the alternative and what is the typical pitfalls when we do clustering together}; (2) performing adaptive clustering that achieves higher recall and overall performance without introducing any extra parameters.
we propose a more fine-grained grouping strategy, which first explicitly addresses the expansion of existing seed attributes, then discovers new potential attributes, as described below.

\paragraph{Adaptive Expansion on Existing Attributes}
We borrow the concept from few-shot learning, and regard each existing seed attribute set as a support set.
The distance between each candidate span $c_i$ and each support set $\mathcal{S}_j$ is measured by $\mathcal{D}$, which is the averaged cosine distance between the candidate and each seed values, as in Eq~\eqref{eq:cluster-adaptive-dist}.
A candidate $c_i$ is added to an attribute $j$ if $\mathcal{D}(c_i, \mathcal{S}_j) < \mathbf{t}_j$, where $\mathbf{t}_j$ is a threshold calculated adaptively based on its support set, as in Eq~\eqref{eq:cluster-adaptive-th}. Particularly, $\delta \in (0, 1]$ is a hyperparameter to relax the threshold that can be tuned on the development set.
\begin{align}
    \mathcal{D}(c_i, \mathcal{S}_j) &= \frac{1}{|\mathcal{S}_j|} \sum_{s_k \in \mathcal{S}_j} \text{cosine}(c_i, s_k) \label{eq:cluster-adaptive-dist} \\
    \mathbf{t}_j = \delta &\cdot \frac{1}{|\mathcal{S}_j|^2} \sum_{s_u, s_v \in \mathcal{S}_j} \text{cosine}(s_u, s_v) \label{eq:cluster-adaptive-th}
\end{align}

\paragraph{More Attribute Coverage}
For remaining candidate spans, more clusters are mined to increase coverage primarily for potential new attributes. We also resort to off-the-shelf DBSCAN that can automatically discover clusters and distinguish noises based on the pairwise cosine distance.

The union of clusters from the above two stages serve as the final result of the candidate grouping.

% \section{Supervision for New Attributes}
\section{Implicit Signals for New Discovery}
\label{sec:new-attr}
% \chenwei{when we introduce intuitions backened with propoer stats in section 2, likely we can make a stronger claim here (sth like ``the majority of the phrases lack proper direct supervision'' ->}
% As Section~\ref{subsec:attribute-generation}-\ref{subsec:attribute-grouping} introduce our entire three-stage framework, we further propose to leverage more implicit signals as weak regularization in the representation learning stage, as described below.
% \chenwei{nicer when the main illustrative figure captured all intuitions below using simple examples}

% Section~\ref{subsec:attribute-repr} introduces the \textbf{supervised} contrastive learning that directly utilizes the seed supervision.
% To maximize the information gain from the weakly-supervised training, we later introduce weak regularization in Section~\ref{subsec:attribute-reg}, where we further formulate our representation learning from two other unique perspectives: \textbf{self-supervised} signals from designed heuristics, and \textbf{unsupervised} signals through neural topic modeling.

Since the seed set only provides semantic signals regarding seed attributes, the majority of candidate spans lack proper supervision, as most of them are absent from the seed set,
% \chenwei{refer to the number in the table}
especially for those new attributes that have no direct supervision during representation learning.
Therefore, it is desirable to exploit additional \textbf{implicit signals} towards more new-attribute-friendly embedding space, and we propose novel methods to tackle the challenge by fully leveraging product context through \textbf{self-supervised} and \textbf{unsupervised regularization}.

\subsection{Self-Supervised Contrastive Learning}
\label{subsec:self-supervised}

% While the supervised contrastive learning on the seed set can be straightforward and has been well investigated by previous work, the majority of the text corpus however does not participate in the training directly, as only a small portion of text is covered by the seed set. Therefore, the training does not utilize more input context other than seeds. Additional signals could be potentially mined from the context that benefit the representation learning.

To utilize the product context, we formulate a self-supervised contrastive heuristic similar to skip-gram in word2vec \cite{word2vec}. We regard each bullet point as a window: pushing two candidate spans within the same window (same bullet point) to have closer representation than two spans not in the same window (different bullet points of a product). It is based on the general observation that different bullet points usually discuss different product perspectives, but within each point, similar attributes or topics are usually mentioned. Though noisy, useful semantic signals could still be revealed given enough corpus, similar to the skip-gram training.

Let $I^b$ be all candidate span indices in bullet points, $P^b(i)$ be the indices of positive spans within the same bullet point as $i$, $N^b(i)$ be the corresponding negative spans from different bullet points of the same product. 
The self-supervised contrastive loss is denoted as:
% \begin{align}
% % \label{eq:attr-self-sup}
%     &\mathcal{L}^{sf} = \nonumber \\ &\sum_{i \in I^b} \frac{-1}{|P^b(i)|} \sum_{p \in P^b(i)} \log \frac{e^{(g_i \cdot g_p / \tau)}}{\sum_{j \in A(i) \setminus P^b(i)} e^{(g_i \cdot g_j / \tau)}} \nonumber
% \end{align}
\begin{align}
% \label{eq:attr-self-sup}
    \mathcal{L}^{ss} = \sum_{i \in I^b} \frac{-1}{|P^b(i)|} \sum_{p \in P^b(i)} \log \frac{e^{(g_i \cdot g_p / \tau)}}{\sum_{j \in N^b(i)} e^{(g_i \cdot g_j / \tau)}} \nonumber
\end{align}
We regard $\mathcal{L}^{ss}$ as a form of regularization, assigning a small coefficient during training. The final loss is described as in Eq~\eqref{eq:attr-final-loss}.

\subsection{Unsupervised Latent Attributes}
\label{subsec:unsupervised}
% \chenwei{to me, it would be more ideal that we start with why sec 5.1 does not suffice? e.g. captures pairwise/triplewise value-level relations but does not capture type-level distribution information as well as type-value distribution info}
More useful signals could still be revealed from product context in addition to the bullet point heuristic.
Inspired from topic modeling, e.g. Latent Dirichlet Allocation (LDA) \cite{lda}, a classic generative method that discovers latent topics unsupervisely from bag-of-words documents, here we propose a formulation of latent attributes to regulate the embedding space, providing implicit signals based on the semantic distribution of corpus, especially beneficial to new attribute discovery that has no direct supervision.
We adapt the neural LDA work from \citet{miao-neural-lda,dieng-etal-2020-topic}, and regard topics as attributes in our setting. The main idea is that each product can be rendered as a composition of spans (equivalently, bag-of-spans) generated from different latent attributes based on the following two distributions.

% Apart from the self-supervised heuristic, we further propose another unsupervised regularization by adapting topic modeling in the training.
% Latent Dirichlet Allocation (LDA) \cite{lda}, as a classic generative topic modeling method, operates on bag-of-words document representation and discovers latent topics unsupervisely. Various neural LDA techniques have since been studied, and we adapt the work from \citet{miao-neural-lda,dieng-etal-2020-topic} to regulate latent attributes from corpus.

% We regard latent topics as latent attributes in our setting, and the main idea is that each product can be rendered as a composition of spans (equivalently, bag-of-spans) generated from different latent attributes based on the following two distributions.

\paragraph{Product-to-Attribute Distribution}
Given the context of a product, the model predicts a distribution over $K$ latent attributes, where $K$ is a hyperparameter. Latent attributes of higher probabilities play a larger role in a product's semantics.
Since learning the true distribution is intractable, variational inference is applied such that we posit the distribution family to be multivariate Gaussian with diagonal covariance matrix, and fix the prior distribution as standard Gaussian \cite{dieng-etal-2020-topic}. Hence, the posterior Product-to-Attribute distribution can be obtained by simply predicting the mean and variance of multivariate Gaussian. Let $p$ represent a product, $\mathbf{h}^p$ be its context representation, $\mu_{k}^p / \sigma_{k}^p$ be its mean/variance for the latent attribute $k$ predicted by the model. A sampled probability of attribute $k$ for product $p$ can be denoted as $\alpha_{k}^p$:
\begin{align}
\label{eq:prod-to-attr}
    \mu_{k}^p / \sigma_{k}^p &= W^{\mu / \sigma}_k \cdot \mathbf{h}^p \\
    \widetilde{\alpha}_{k}^p &\sim \mathcal{N}(\mu_{k}^p,\, \sigma_{k}^p) \\
    \alpha_{k}^p &= \text{softmax}\, (\widetilde{\alpha}_{k}^p) \;\vert_{k=1}^K
\end{align}
$W^{\mu / \sigma}_{k}$ is a learned parameter to predict mean and variance. For $\mathbf{h}_p$, we use the averaged CLS representation of its product title and all bullet points.

\paragraph{Attribute-to-Span Distribution}
For each latent attribute, the model also learns a distribution over candidate spans; spans of high probabilities are the representatives of this attribute. Following \citet{dieng-etal-2020-topic}, rather than building an explicit distribution, the model instead simply learns an attribute embedding, so that the distribution can be obtained by measuring the similarity of the attribute embedding and span embeddings.
Let $h_k$ be the k'th attribute embedding learned by the model, $g_c$ be the representation of a candidate span $c$, and $\mathcal{C}$ be all unique candidate spans from all products in a training batch. The distribution of an attribute $k$ over candidates $\mathcal{C}$ can be denoted as:
\begin{align}
\label{eq:attr-to-span}
    \beta_{kc} = \text{softmax}\, (h_k \cdot g_c) \;\vert_{c \in \mathcal{C}}
\end{align}

\paragraph{Optimization}
Given the above two distributions for a product $p$, the model can easily get the \textbf{Product-to-Span} distribution $\mathcal{P}(c|p)$ by marginalizing out the latent attributes, as in Eq~\eqref{eq:prod-to-span}, which can then be used to optimize a reconstruction objective, such that spans actually appeared in product $p$ should have higher probability than those who do not. Let $V(p)$ be the candidate spans in a product $p$, $m$ be the total number of products. The unsupervised reconstruction loss $\mathcal{L}^{un}$ can be estimated by evidence lower bound (ELBO) as:
\begin{align}
    & \mathcal{P}(c | p) = \sum_{k=1}^K \alpha_{k}^p \cdot \beta_{kc} \label{eq:prod-to-span} \\
    \mathcal{L}^{un} &= -\sum_{p=1}^{m} \big( \sum_{c' \in V(p)} \log \mathcal{P}(c' | p) + \text{KL}(\widetilde{\alpha}^p \Vert \hat{\alpha}) \big) \nonumber \\
    \mathcal{L} &= \mathcal{L}^{su} + \lambda^{ss} \cdot \mathcal{L}^{ss} + \lambda^{un} \cdot \mathcal{L}^{un} \label{eq:attr-final-loss}
\end{align}
where $\hat{\alpha}$ is the fixed standard Gaussian (prior Product-to-Attribute distribution). The first term of $\mathcal{L}^{un}$ is the log-likelihood to encourage higher probability for actually appeared candidate spans in a product, and the second KL-divergence term regularizes the posterior attribute distribution $\widetilde{\alpha}_p$ to be close to the standard Gaussian $\hat{\alpha}$.

The final loss $\mathcal{L}$ during representation learning is constituted by three losses; $\lambda^{ss}$ and $\lambda^{un}$ are hyperparameters that control the regularization strength.
% \chenwei{ever considered having 3.3 after 4.2?}
\section{Experiments}
\label{sec:experiments}

\begin{table*}[tbp!]
\centering
\resizebox{0.96\textwidth}{!}{
\begin{tabular}{p{2.3cm}l|ccccccccccc}
\toprule
&& \multicolumn{5}{c}{Exact Match} && \multicolumn{5}{c}{Partial Match} \\
\cmidrule{3-7} \cmidrule{9-13}
&& Jaccard & ARI & NMI & Recall & \bf F1 && Jaccard & ARI & NMI & Recall & \bf F1 \\
\midrule
\multirow{2}{*}{\shortstack[c]{Closed-World\\{\it (Tagging)}}} & \tt Tx-CRF & \bf 92.5 & \bf 95.4 & \bf 95.8 & 20.0 & 32.8 && \bf 78.2 & \bf 85.3 & \bf 86.7 & 30.5 & 44.2 \\
& \tt SU-OpenTag & 70.1 & 78.8 & 87.1 & \bf 22.1 & \bf 34.5 && 61.7 & 72.6 & 79.5 & \bf 34.7 & \bf 46.6 \\
\midrule
\multirow{2}{*}{\shortstack[l]{\;Open-World\\ \;{\it \;\;(Segment)}}} & \tt OA-Mine & \bf 63.5 & 74.4 & 78.8 & 25.3 & 36.9 && 48.8 & 60.9 & 64.9 & 40.5 & 46.7 \\
& \tt Amacer\textsuperscript{*} & \bf 69.9 & \bf 78.0 & \bf 84.1 & \bf 29.0 & \bf 41.7 && \bf 58.4 & \bf 68.8 & \bf 73.7 & \bf 47.8 & \bf 54.9 \\
\midrule
\multirow{5}{*}{\shortstack[l]{\;Open-World\\ \;{\it\;\;\;(Syntax)}}} & \tt DBSCAN & 22.4 & 29.8 & 69.5 & 17.3 & 23.6 && 20.6 & 24.7 & 60.7 & 26.9 & 30.3\\
& \tt DBSCAN+AE & 32.8 & 41.8 & 61.2 & 30.3 & 35.9 && 25.1 & 30.1 & 47.1 & 50.5 & 40.7 \\
& \tt OA-Mine\textsuperscript{*} & 55.8 & 68.2 & 73.6 & 30.8 & 41.1 && 40.6 & 52.0 & 57.2 & 50.1 & 49.8 \\
& \tt Amacer\textsuperscript{-R} & 58.3 & 69.6 & 79.2 & 35.5 & 46.3 && 46.3 & 57.6 & 65.8 & \bf 57.7 & 56.9 \\
& \tt Amacer & \bf 67.2 & \bf 76.9 & \bf 84.0 & \bf 35.7 & \bf \underline{47.6} && \bf 52.7 & \bf 63.8 & \bf 70.4 & 57.1 & \bf \underline{59.1} \\
\bottomrule
\end{tabular}}
\caption{Evaluation results on the test set of our new dataset \textsc{WoaM}, with F1 being the overall evaluation metric. See Section~\ref{sec:experiments} for detailed specifications of model settings and evaluation metrics. Each number is the macro-average across all product categories. Models with lower Recall tend to have higher Jaccard/ARI/NMI scores, as they produce fewer (and easier) attribute clusters of higher purity. The best performance by both Exact/Partial-F1 is the underlined score achieved by our approach \texttt{Amacer} (statistically significant from t-test $> 95\%$ confidence).}
% \jingbo{what does that mean for the underline in the table?}}
\label{tab:result}
\vspace{-1.5ex}
\end{table*}

Experiments are conducted on our dataset in multiple model settings, including various baselines. Three different types of models are examined based on how attribute spans are obtained:

(1) Closed-world models based on sequence-tagging that extract spans upon predicted BIO tags of existing attributes, which do not support new attribute discovery natively. Two models are experimented:
\textbf{\texttt{Tx-CRF}}, a generic Transformers-CRF tagging model; 
\textbf{\texttt{SU-OpenTag}} \cite{su-opentag}, a popular tagging-based attribute extraction model.

(2) Open-world models that rely on sentence segmentation to obtain candidate spans. We use the code released from OA-Mine to obtain all text segments for our dataset. Two settings are included:
\textbf{\texttt{OA-Mine}} \cite{oamine}; \textbf{\texttt{Amacer\textsuperscript{*}}}, a stripped version of our approach removing regularization and directly taking segments as candidates.

(3) Open-world models that employ our syntax-based candidate generation (\textsection\ref{sec:attribute-generation}). Five settings are included: 
\textbf{\texttt{DBSCAN}} that directly performs DBSCAN clustering without representation learning;
\textbf{\texttt{DBSCAN+AE}} that adds our proposed adaptive expansion (\textsection\ref{sec:seed-attr});
\textbf{\texttt{OA-Mine\textsuperscript{*}}} that substitutes segmentation with our candidate spans;
\textbf{\texttt{Amacer}}, our full proposed approach; and \textbf{\texttt{Amacer\textsuperscript{-R}}} that only utilizes seed supervision without regularization in \textsection\ref{sec:new-attr}.

For candidate span generation, we use spaCy\footnote{\url{https://spacy.io}} to obtain POS tags; a total of 96 valid POS patterns are acquired from product profiles (Section~\ref{sec:attribute-generation}).
The same BERT-Large is used as the encoder for all models. Our detailed hyperparameter settings are provided in Appendix~\ref{apdx:settings}.

\paragraph{Evaluation Metrics}
Standard clustering evaluation metrics are used:
\textbf{Jaccard}, Adjusted Rand Index (\textbf{ARI}), Normalized Mutual Information (\textbf{NMI}), to compare the attribute assignments on gold spans; \textbf{Recall}, to evaluate gold cluster coverage. 
As above metrics are consistent with OA-Mine, the evaluation adopts exact-match on predicted/gold spans. However, it could become over-restrictive as span boundaries can be quite subjective in this open-world setting, losing the information of near-correct predictions. Thus, we also provide a relaxed evaluation that allows partial-match on spans, such that a predicted span is considered an attribute value if more than half of the span falls into a gold value.

To assess the overall performance of a model, we roughly regard the averaged number of Jaccard, ARI and NMI as \textit{pseudo precision}, and derive a single pseudo-\textbf{F1} score based on the clustering \textit{precision} and \textit{recall}, serving as the main evaluation metric of each approach.

% \subsection{Results}
% \label{subsec:result}
% \chenwei{given separated clustering on seed values from existing attributes vs. new attribute values, are we able to report results separately as well (on existing attributes/ on new attributes) and any specific metric we can report to further show our advantages? (e.g. \% of clusters corresponding to new attributes? try to develop specific claims that people can remember, and help people understand where we are in tackling the task)}

\paragraph{Results}
Table~\ref{tab:result} shows the evaluation results by all model settings.
Our full proposed approach \texttt{Amacer} surpasses both \texttt{SU-OpenTag} and \texttt{OA-Mine} by a large margin (10+ Exact/Partial-F1), achieving the best performance on this task. Further observations and ablation study can be obtained as below.

• \textbf{Open-world models identify more attributes than closed-world models.}
The two tagging-based models underperform OA-Mine-based models and our Amacer-based models, with noticeably lower recall. It can be attributed to two factors.
First, as all spans are obtained through tagging learned solely from the seed set, they lack the ability to accept more diverse attribute values not covered in training, not being able to generalize well under limited supervision. Second, new attributes are left untouched, unlike the open-world counterparts.

• \textbf{Adaptive expansion on seed attribute types is effective for candidate grouping.}
By simply comparing \texttt{DBSCAN} with \texttt{DBSCAN+AE}, adaptive expansion is shown greatly improving the recall by 13-23\% and overall performance by 10+\%. On a side note, there is still a huge gap between \texttt{DBSCAN+AE} and \texttt{Amacer}, demonstrating the necessity to refine embedding space by representation learning.
% the only difference between \texttt{DBSCAN} and \texttt{2SC} is that \texttt{2SC} additionally applies the proposed adaptive clustering on seed attributes, directly bringing 10+ exact/partial F1 gain, especially with a much higher recall (+23\% on partial-match), which validates our strategy to utilize seed sets to increase seed attribute prediction coverage.

• \textbf{Syntax-oriented generation obtains candidate spans of higher quality than segmentation.}
Both \texttt{OA-Mine\textsuperscript{*}} and \texttt{Amacer\textsuperscript{-R}} that apply syntax-oriented candidates outperform their segmentation-based counterparts \texttt{OA-Mine} and \texttt{Amacer\textsuperscript{*}}, especially for exact-match that brings a gap of 4+ F1.
% by comparing \texttt{OA-Mine} with \texttt{OA-Mine\textsuperscript{*}}, or \texttt{Amacer\textsuperscript{-}} with \texttt{Amacer}, we can see that simply substituting segmentation with our proposed filtered n-grams as candidates improves the overall performance by up to 4.6 Exact-F1. 
Notably, our generation step takes under 10 minutes to process each category on CPUs, while the segmentation requires several hours on a GPU. Qualitatively, we found that the segmentation often over-divides sentences, yielding many noisy and incomplete phrases.

% given the substantial gap between \texttt{2SC} and \texttt{Amacer}, it is clear that performing attribute-aware representation learning is crucial in this pipeline.
• \textbf{Seed supervision is more efficiently utilized by in-batch negative contrastive loss.}
Compared to the triplet loss and regression loss adopted in \texttt{OA-Mine\textsuperscript{*}}, the in-batch loss is not only simpler but also improves 5+ F1 in this task. We found the regression loss that pushes cosine similarity to 1/-1 for pos/neg pairs can be too harsh for the embedding space, as certain attribute types are indeed more related and not completely independent.
% Although \texttt{OA-Mine} also employs supervised contrastive loss, our analysis identifies two factors that reduce its performance. First, \texttt{OA-Mine} applies a regression loss in training as well that pushes cosine similarity to 1/-1 for pos/neg pairs, which can be too harsh, as certain attributes are indeed more related than others. 
% For follow-up works, we recommend to adopt the in-batch negative contrastive loss as in this work due to its empirical superiority.
% Second, the grouping stage in \texttt{OA-Mine} is not as effective as \texttt{Amacer}. In fact, we can get a direct boost of 3.5 Partial-F1 on \texttt{OA-Mine\textsuperscript{*}} by simply replacing its grouping with our proposed \textsection\ref{subsec:attribute-grouping}.

• \textbf{Regularization (\textsection\ref{sec:new-attr}) is able to bring additional semantic signals} useful to shape the attribute-aware embedding space, as shown by the 2.2 Partial F1 improvement of \texttt{Amacer} upon \texttt{Amacer\textsuperscript{-R}}, where the unsupervised latent attribute formulation contributes around 70\% improvement. We provide further quantitative and qualitative insights in Section~\ref{sec:quant-analysis}-\ref{sec:qual-analysis}.
% our ablation study shows that around $76\%$ improvement of \texttt{Amacer\textsuperscript{+}} upon \texttt{Amacer} comes from the unsupervised topic modeling. More qualitative analysis regarding \textsection\ref{subsec:attr-unsupervised} is provided in Section~\ref{sec:qual-analysis}.

\section{Quantitative Analysis}
\label{sec:quant-analysis}

\begin{table}[tbp!]
\centering
\resizebox{0.82\columnwidth}{!}{
\begin{tabular}{l|ccc}
\toprule
& Seed / New & Title / BP & Gold\\
\midrule
\tt OA-Mine\textsuperscript{*} & 51.2 / 24.6 & 56.6 / 49.0 & 61.2 \\
\tt Amacer\textsuperscript{-R} & 64.5 / 39.8 & 61.2 / 56.9 & 69.8 \\
\tt Amacer & \bf 66.0 / \bf 46.2 & \bf 61.5 / \bf 59.3 & \bf 71.9 \\
\bottomrule
\end{tabular}}
\caption{Decomposed evaluation (Partial-F1) by: seed attribute types only (Seed) / new attribute types only (New); product titles only (Title) / bullet points only (BP). Gold shows the result by taking gold values directly as candidate spans. Full metrics are provided in Table~\ref{tab:result-seed-new}-\ref{tab:result-gold} (Appendix~\ref{adpx:quant}).}
\label{tab:analysis-cmp}
\vspace{-1ex}
\end{table}

\begin{table}[tbp!]
\centering
\resizebox{0.98\columnwidth}{!}{
\begin{tabular}{l|ccccccc}
\toprule
& \multicolumn{3}{c}{Span (Exact)} && \multicolumn{3}{c}{Span (Partial)} \\
\cmidrule{2-4} \cmidrule{6-8}
& P & R & F && P & R & F \\
\midrule
\tt OA-Mine\textsuperscript{*} & 31.0 & 38.3 & 34.2 && 52.8 & 64.8 & 58.1 \\
\tt Amacer\textsuperscript{-R} & 27.8 & \bf 41.9 & 33.4 && 46.7 & \bf 70.3 & 56.1 \\
\tt Amacer & \bf 33.5 & 40.5 & \bf 36.4 && \bf 54.9 & 65.5 & \bf 59.3 \\
\bottomrule
\end{tabular}}
\caption{Evaluation of precision/recall/F1 (P/R/F) on the final extracted spans against gold values by exact/partial-match, regardless of the attribute types.}
\label{tab:analysis-me}
\vspace{-1.5ex}
\end{table}

To quantify the unique challenges of this task, we decompose the evaluation to examine two perspectives specifically:
\begin{itemize}[noitemsep,nolistsep,leftmargin=*]
\item Performance on new attribute types (only open-world evaluation) compared to seed types (only closed-world evaluation).
\item Performance on attribute values in bullet points compared to titles.
\end{itemize}

% \textit{New Attributes}:
Table~\ref{tab:analysis-cmp} shows that all models suffer performance degradation on new attribute types unseen in training, comparing with those existing seed types, which corroborates the expectation that open-world discovery remains a tough challenge owing to no direct supervision. It is noteworthy that our approach brings significant improvement on new attributes; especially, our proposed regularization in \texttt{Amacer} boosts performance on existing types by relatively 2.3\% upon Amacer\textsuperscript{-R}, while the improvement on new types is 16.1\%, which fulfills our motivation to provide semantic supervision for those new attributes. Compared to \texttt{OA-Mine\textsuperscript{*}}, our approach exhibits smaller relative gap between existing and new types, discovering 39\% new types (Recall in Table~\ref{tab:result-seed-new}).

% Noticeably, adding our proposed weak regularization \texttt{Amacer\textsuperscript{+}} reduces the performance gap by 4.9 Partial-F1, discovering more open-world attributes, thus demonstrates the advantage of learning more implicit signals from corpus.

For more traits of our corpus, all models struggle to keep up the performance on bullet points compared to titles, showing that they are indeed harder to extract from due to their characteristics (Table~\ref{tab:dataset-brief}\&\ref{tab:title-bp}).
Interestingly, our proposed regularization is also able to reduce the gap from 4.3 to 2.2 Partial-F1, which can be credited to both self-supervised heuristic and unsupervised latent attributes, as they both leverage the product context mainly from bullet points.
% \texttt{Amacer\textsuperscript{+}} is also able to reduce the gap from 4.3 to 2.2 Partial-F1, while \texttt{OA-Mine\textsuperscript{*}}, originally designed only for titles, shows a huge gap of 9.3 Partial-F1.

To detach the impact of candidate generation, we provide additional views to assess the representation learning and grouping performance.
The last column of Table~\ref{tab:analysis-cmp} shows evaluation by using gold values as candidate spans directly. It clearly strengthens the advantage of our proposed representation learning methods, as \texttt{Amacer} outperforms \texttt{OA-Mine\textsuperscript{*}} by 10+ Partial-F1.

Table~\ref{tab:analysis-me} further evaluates span extraction of predicted values against gold values.
All models are shown quite low Exact-F1 scores ($<37$) and low precision ($<34$), leaving room for future improvement to extract more correct candidate spans under limited supervision.
% By using gold spans as candidate spans directly (right table), the performance from all models gets boosted as expected; still, \texttt{Amacer\textsuperscript{+}} keeps robust improvement.

\section{Qualitative Analysis}
\label{sec:qual-analysis}

% We examine the model predictions by three perspectives as below. Overall, we found that our approach performs quite well for existing seed attributes, but understandably, still requires future research to strengthen the new attribute discovery that provides no direct signals during training.

\begin{table*}[htbp!]
\centering
\resizebox{\textwidth}{!}{
\begin{tabular}{l}
\toprule
\makecell[tl]{Our \colorbox{cyan}{English Afternoon tea} combines Keemun tea from the  \textit{\textbf{\underline{\colorbox{cyan}{Anhui province}}}} in \textit{\textbf{\underline{\colorbox{cyan}{China}}}} with  \colorbox{cyan}{Ceylon tea} from \textit{\textbf{\underline{\colorbox{orange}{Sri Lanka}}}}. \\  \colorbox{cyan}{Keemun teas} are \colorbox{green}{smooth} and slightly \colorbox{green}{sweet in taste}, while \colorbox{cyan}{Ceylon teas} are \colorbox{green}{crisp and refreshing}.} \\
\midrule
\colorbox{orange}{Wormwood} (\colorbox{orange}{Artemisia absinthium}) is a \colorbox{green}{bitter herb} found in \textit{\textbf{\underline{\colorbox{orange}{Eurasia}}}}, \textit{\textbf{\underline{\colorbox{orange}{North Africa}}}}, and \textit{\textbf{\underline{\colorbox{orange}{North America}}}}. \\
\bottomrule
\end{tabular}}
\caption{Examples of extracted spans by \texttt{Amacer} on two bullet point description; the colors of spans represent three predicted seed attribute types: \textcolor{cyan}{\textit{drink type}}, \textcolor{orange}{\textit{ingredient}}, \textcolor{green}{\textit{flavor profile}}. According to the gold annotation, the following spans should belong to a separate attribute cluster (marked as a new non-seed attribute type ``\textit{region of origin}''): \textit{\textbf{Anhui province}}, \textit{\textbf{China}}, \textit{\textbf{Sri Lanka}}, \textit{\textbf{Eurasia}}, \textit{\textbf{North Africa}}, \textit{\textbf{North America}}. The model mistakenly predicts them as two existing attributes, showing that open-world attribute discovery remains a tough challenge to be solved under this task setting. On the other hand, it is still encouraging to see these spans being extracted and recognized as certain attributes, since the model has not seen any location-specific attributes directly from the seed set.}
\label{tab:result-example-new}
% \vspace{-1.5ex}
\end{table*}

\begin{table*}[tbp!]
\centering
\resizebox{\textwidth}{!}{
\begin{tabular}{llllll}
\multicolumn{6}{c}{6 Selected Learned \textit{Latent Attributes} by Each Column} \\
\midrule
\makecell[tl]{living room \\ navy love seats \\ tufted sofa \\ upholstered loveseat \\ velvet sofa} &
\makecell[tl]{orange \\ purple clear \\ virtually invisible \\ brown hue \\ warm neural} &
\makecell[tl]{oolong \\ black tea \\ green tea \\ ti kuan yin oolong \\ herbal tea} &
\makecell[tl]{no synthetic dyes \\ premium ingredients \\ artificial ingredients \\ vegetarian \\ vegan and gluton free} &
\makecell[tl]{vitamin d3 \\ kids vitamin c \\ vitamin b12 \\ amino acids \\ folic acid} &
\makecell[tl]{moto g pure \\ 12 \\ apple iphone \\ nokia x100 \\ galaxy s21 fe} \\
\midrule
\end{tabular}}
\caption{Examples of several learned latent attributes, with top candidate spans from corpus at each column (high-probability spans in each Attribute-to-Span distribution). These learned latent attributes can represent certain concepts and provide additional semantic signals during representation learning, especially for new attributes.}
% \chenwei{show by pt? otherwise i feel unnecessarily that we define abbreviation for each pt in the paper (can be moved to appendix)}
\label{tab:result-example-topic}
\vspace{-1ex}
\end{table*}

\quad \textbf{Seed Attributes}: our approach performs generally well on seed attribute types. Table~\ref{tab:seed-example-brief} shows examples of discovered new values on a seed type \textit{Flavor Profile} (also see Table~\ref{tab:result-example}). 
\texttt{Amacer} is able to extract sensible and diverse expressions, given only 6 seed values as supervision.
Each proposed component makes evident contribution: the candidate generation can capture unseen long-tail spans, such as \textit{floral with honey notes}, \textit{delicate zesty}, while the representation learning and grouping together are effective recognizing similar attribute values.
Nearly 80 new flavor values are identified on our test set, expanding its vocabulary by 12 times.

\begin{table}[htb!]
\centering
\resizebox{0.85\columnwidth}{!}{
\begin{tabular}{ll}
\multicolumn{2}{c}{\it \large Flavor Profile} \\
\cmidrule{1-2}
\textbf{Seed} (6) & \textbf{Extracted} (80+) \\
\toprule
\makecell[tl]{sweet \\ sweetened \\ unsweetened \\ sour \\ bitter \\ fruity} &
\makecell[tl]{ nutty \\ floral with honey notes \\ earthy \\ tangy and fruity \\ sweet and savory spice flavors \\ smokiness \\ delicate zesty \\ refreshingly tart herbal } \\
\end{tabular}}
\caption{Sampled predictions on TEA products of the seed attribute \textit{Flavor Profile} capturing diverse new values. Full examples are provided in Table~\ref{tab:result-example}.}
\label{tab:seed-example-brief}
\vspace{-1ex}
\end{table}

\textbf{New Attributes}: it is inevitably difficult to discover values of new types, as models possess little prior knowledge as regards. For error analysis, we found that for most of these new types, their values are either absent in the predictions, or grouped as other existing attributes mistakenly. Table~\ref{tab:result-example-new} shows an example of the latter case; however, it is still encouraging that these new values are extracted and recognized as certain attributes, rather than being neglected by the model, which partially achieves the open-world discovery objective.
% Indeed, more future research is needed to resolve this open-world expectation.

\textbf{Latent Attributes}: Table~\ref{tab:result-example-topic} shows examples of learned latent attributes resulted by contrastive loss and topic modeling. They resemble certain ``concepts'' that regulate towards more attribute-friendly embedding space.
However, we also observe that certain learned attributes are repetitive, such that their attribute embeddings have high cosine similarity. This behavior aligns with the previously discovered issue known as \textit{topic collapsing} \cite{srivastava2017autoencoding}, leading to deficient discovery. We do not particularly address it in this work, and leave it for future research.

% \begin{table}[tbp!]
% \centering
% \resizebox{\columnwidth}{!}{
% \begin{tabular}{l|ccccc}
% \toprule
% & \multicolumn{2}{c}{Exact Match} && \multicolumn{2}{c}{Partial Match} \\
% \cmidrule{2-3} \cmidrule{5-6}
% & Seed / New & Title / BP && Seed / New & Title / BP \\
% \midrule
% \tt OA-Mine\textsuperscript{*} & 43.4 / 21.6 & 48.4 / 39.6 && 51.2 / 24.6 & 56.6 / 49.0 \\
% \tt Amacer\textsuperscript{-R} & 53.9 / 31.0 & 50.9 / 45.3 && 64.5 / 39.8 & 61.2 / 56.9 \\
% \tt Amacer & \bf 54.8 / \bf 36.2 & \bf 51.5 / \bf 46.8 && \bf 66.0 / \bf 46.2 & \bf 61.5 / \bf 59.3 \\
% \bottomrule
% \end{tabular}}
% \caption{Decomposed evaluation (F1) on: seed attribute types only (Seed) / new attribute types only (New); product titles only (Title) / bullet points only (BP). Full metrics are provided in Table~\ref{tab:result-seed-new}-\ref{tab:result-title-bp}.}
% \label{tab:analysis-cmp}
% \vspace{-1ex}
% \end{table}
% \chenwei{i don't see we dicussing details about the ``39\%'' mentioned in the abstract}
\section{Conclusion}
\label{sec:conclusion}

In this work, we present a new task setting as a practical solution to mine open-world attributes without extensive human intervention. A new dataset is created accordingly, and our proposed approach is designed for light supervision, especially by utilizing a high-quality seed set, as well as exploiting self-supervised and unsupervised semantic signals from the context. Empirical results show that our approach effectively improves discovery upon baselines on both existing and new attribute types.

\section{Limitations}
\label{sec:limitations}

The scope of our approach is intended for our specific task setting, which is proposed as a practical solution to mine open-world attributes without heavy supervision, and has not been studied previously.
Our approach does require an external dependency of a POS tagger, and assumes high POS tagging quality on English. Thankfully, there are POS tools publicly available with high performance, and are quite robust against domain shift, mostly fulfilling the assumption.

Our current candidate generation that utilizes syntax-oriented patterns does not check the semantics, which can be another limitation. It introduces noisy spans in the process, such as \textit{``supports joint health \& overall''} (in Table~\ref{tab:result-example}). Future works could consider combining syntax with semantics to alleviate noisy spans.

\bibliography{references}
\bibliographystyle{acl_natbib}

\clearpage
\appendix
\section{Previous Work}
\label{apdx:prev}

As the most related previous work to our proposed task setting is OA-Mine \cite{oamine}, we found that their released dataset is not ideal nor practical to serve as the testbed for this setting, due to three drawbacks:

\begin{itemize}[noitemsep,nolistsep,leftmargin=*]
    \item The seed attribute set is too sparse: there are only five seed values provided for each attribute type, leading to insufficient attribute extraction and discovery.
    \item The seed attributes can be quite noisy; especially, certain values appear under multiple attribute types, presenting noise and ambiguity to the model training (example shown in Table~\ref{tab:oamine-example}).
    \item The corpus only consists of product titles, and lacks the full product description taxonomy such as bullet points, which can provide richer information regarding attributes and also require stronger inference capability. Detailed statistics of bullet point description compared to titles are provided in Table~\ref{tab:title-bp}.
\end{itemize}

Our dataset explicitly addresses above issues, and is constructed to provide higher quality and richer context, as introduced in Section~\ref{sec:data}.

\begin{table}[h]
\centering
\resizebox{0.9\columnwidth}{!}{
\begin{tabular}{l|ccccc}
\toprule
& Tok & Cand & Seed & Gold & Type (New) \\
\midrule
\tt TT & 20.1 & 7.3 & 2.9 & 5.7 & 46 (28.3\%) \\
\tt BP & 26.6 & 8.8 & 1.2 & 3.6 & 65 (43.1\%) \\
\bottomrule
\end{tabular}}
\caption{Statistics of our dataset \textsc{WoaM} that show more comparison between product titles (\texttt{TT}) and bullet point description (\texttt{BP}). \underline{Tok} is the averaged number of tokens per sequence; \underline{Cand} is the averaged number of generated candidates described in Section~\ref{sec:attribute-generation}. \underline{Seed} is the averaged occurrences of seed values per sequence, and \underline{Gold} is the averaged occurrences of gold values in the test set. \underline{Type} denotes the total number of attribute types in the test set, with parentheses indicating the ratio of new types that do not exist in the seed attribute set.}
\label{tab:title-bp}
% \vspace{-1.5ex}
\end{table}

\begin{table*}[bh]
\centering
\resizebox{0.74\textwidth}{!}{
\begin{tabular}{lll}
\toprule
\multicolumn{1}{c}{Seed Attribute Type} && \multicolumn{1}{c}{Seed Attribute Values} \\
\midrule
material feature && \it \colorbox{cyan}{organic},\; gmo free,\; \colorbox{green}{kosher},\; caffeine free,\; \colorbox{orange}{gluten free} \\
specialty && \it \colorbox{cyan}{organic},\; natural,\; herbal,\; caffeine free,\; \colorbox{green}{kosher} \\
special ingredients && \it \colorbox{cyan}{organic},\; \colorbox{green}{kosher},\; \colorbox{orange}{gluten free},\; matcha,\; cinnamon \\
diet type && \it \colorbox{orange}{gluten free},\; \colorbox{green}{kosher},\; vegan,\; paleo,\; halal \\
\bottomrule
\end{tabular}}
\caption{An example of seed attributes for TEA products from the dataset released by OA-Mine \cite{oamine}. The provided seed attributes can be quite ambiguous, with many overlapping values in between. As this dataset is constructed in a distant-supervised way, the sub-optimal quality can hinder the model training to discriminate on different attributes. Our seed set adopts a hybrid approach combining data-driven and human curation, producing a practical and higher-quality attribute extraction.}
\label{tab:oamine-example}
% \vspace{-1.5ex}
\end{table*}

\section{Dataset}
\label{apdx:dataset}

\begin{table*}[tbp!]
\centering
\resizebox{\textwidth}{!}{
\begin{tabular}{l|cccccccccccccc}
\toprule
& \multicolumn{5}{c}{Raw Text Corpus} && \multicolumn{3}{c}{Seed Attributes} && \multicolumn{3}{c}{Test Set Attributes} \\
\cmidrule{2-6} \cmidrule{8-10} \cmidrule{12-14}
& TRN & DEV & TST & BP & Toks && Types & Mdn/Avg & Occ && Types (New) & Values (New) & Occ \\
\midrule
% \tt Z-22 & \\
% \midrule
% \midrule
\sc WoaM & 209662 & 4647 & 1425 & 82.8\% & 25.5 && 36 & 9 / 27.0 & 1.5 && 66 (42.4\%) & 3382 (86.9\%) & 3.9 \\
\tt \;-TEA & 49828 & 1094 & 524 & 82.0\% & 22.9 && 14 & 10 / 23.3 & 1.6 && 26 (46.2\%) & 1154 (86.3\%) & 3.7 \\
\tt \;-VIT & 50298 & 1127 & 413 & 82.1\% & 24.1 && 15 & 25 / 37.4 & 1.7 && 22 (31.8\%) & 835 (81.2\%) & 3.5 \\
\tt \;-SOFA & 55655 & 1228 & 240 & 83.8\% & 26.9 && 19 & 9 / 12.8 & 1.3 && 32 (40.6\%) & 775 (92.1\%) & 4.7\\
\tt \;-CASE & 53881 & 1198 & 248 & 83.3\% & 27.5 && 18 & 8 / 15.0 & 1.3 && 30 (40.0\%) & 703 (89.2\%) & 4.6 \\
\bottomrule
\end{tabular}}
\caption{Overall statistics of our created \textsc{WoaM} dataset, with breakdown of each product category:
 \texttt{TEA},  \texttt{VIT} (vitamin),  \texttt{SOFA}, \texttt{CASE} (phone case).
% \texttt{Z-22} is the dataset released by \citet{oamine} for comparison.
\underline{TRN}/\underline{DEV}/\underline{TST}: number of text sequences (titles or bullet points) for the training/development/test set; \underline{BP}: ratio of bullet point sequences; \underline{Toks}: averaged number of tokens per sequence. For the seed set, \underline{Types}: number of seed attribute types, with \underline{Mdn/Avg} being the Median/Averaged number of values per type; \underline{Occ}: averaged occurrences of seed values per sequence. For the test set, \underline{Types/Values}: number of unique attribute types/values by human annotations, with parentheses indicating the ratio of new types/values unseen from the seed set; \underline{Occ}: averaged occurrences of annotated values per sequence.}
\label{tab:dataset}
% \vspace{-1.5ex}
\end{table*}

Full statistics of our new dataset \textsc{WoaM} are provided in Table~\ref{tab:dataset}.
Our dataset is publicly available under the Apache 2.0 License.

\paragraph{Corpus}
Our corpus consists of e-commerce product description from selected product categories, collected under permissions. We do not find concerns regarding privacy issues or discriminatory content.

\paragraph{Product Profiles}
In addition, we also document three detailed issues existed in product profiles that are addressed in our seed set construction: data sparsity, noisy attributes, coarse granularity. Thus, the raw profiles are unable to serve as the full supervision directly for this attribute extraction task.

\begin{itemize}[noitemsep,leftmargin=*]
    \item Our preliminary study shows that 80-90\% human-identified attribute values are missing from the product profiles; along with the missing values, around 40\% identified attribute types are also absent in the profiles, which aligns with the previous observations from \citet{oamine}. The sparsity of product profiles further cultivates our research motivation to enrich the product profiles by discovering new attributes automatically.
    \item Attribute values resided in profiles can be quite noisy, as there are no restrictions on what values that sellers could provide regarding their products. In extreme cases, many irrelevant values may be provided by sellers in efforts to boost their product search performance, which can disrupt the training and make the model insensible. % For instance, \textit{spf 30} appears under many irrelevant attributes of sunscreen products, such as \texttt{COLOR}, \texttt{SIZE}, etc.
    \item Certain attributes may not be used directly due to their coarse granularity. For example, an attribute type \texttt{STYLE} can be too ambiguous for sellers such that it essentially becomes a superset of more fine-grained attribute values including colors, flavors, visual styles, materials, etc.
\end{itemize}

\section{Experimental Settings}
\label{apdx:settings}

For representation learning, BERT-Large \cite{devlin-etal-2019-bert} is adopted as the encoder and we freeze all layers except for the last four layers, allowing for a larger batch size and faster training, which we found performs similar to finetuning the entire BERT. We use a batch size as $128$, learning rate as $2 \times 10^{-5}$, linear-decay learning rate scheduler with warm-up ratio as $0.01$, max gradient clipping norm as $1$.

Other hyperparameters are searched on the development set; in our final \texttt{Amacer} model, we set the temperature $\tau = 0.1$ in the contrastive loss, and the number of latent attributes $K = 50$ (Section~\ref{sec:new-attr}). In the final loss Eq~\eqref{eq:attr-final-loss}, we set $\lambda^{ss} = 0.01$ and $\lambda^{un} = 0.02$, regarding them as weak regularization that mines additional semantic signals.

At the grouping stage, we set the relaxation $\delta = 0.8$ in adaptive expansion Eq~\eqref{eq:cluster-adaptive-th}. For DBSCAN, we use the implementation from sklearn\footnote{\url{https://scikit-learn.org/stable/modules/generated/sklearn.cluster.DBSCAN.html}}, and set eps as $0.05$, min\_samples as $4$.

All training is conducted on a Nvidia Tesla V100 GPU with 32GB memory, and takes around 1 hour to finish each model.

\section{Quantitative Analysis}
\label{adpx:quant}

Full evaluation metrics are provided in
Table~\ref{tab:result-seed-new} and ~\ref{tab:result-title-bp}, in regard to the quantitative analysis in Section~\ref{sec:quant-analysis}.
In particular,
Table~\ref{tab:result-seed-new} separately shows the detailed evaluation results on existing seed attribute types only or on new attribute types only.
Table~\ref{tab:result-title-bp} separately shows the detailed evaluation results on product titles only, or on bullet point description only.

Table~\ref{tab:result-gold} shows the full evaluation metrics when using gold spans as candidate spans directly. Since all resulting spans will be gold values, the evaluation scores are the same for either partial-match or exact-match.

\begin{table*}[tbp!]
\centering
\resizebox{0.91\textwidth}{!}{
\begin{tabular}{cl|ccccccccccc}
\toprule
&& \multicolumn{5}{c}{Exact Match} && \multicolumn{5}{c}{Partial Match} \\
\cmidrule{3-7} \cmidrule{9-13}
&& Jaccard & ARI & NMI & Recall & \bf F1 && Jaccard & ARI & NMI & Recall & \bf F1 \\
\midrule
\multirow{3}{*}{\shortstack[c]{\it Seed}} & \tt OA-Mine\textsuperscript{*} & 50.5 & 64.8 & 73.8 & 33.1 & 43.4 && 36.6 & 50.8 & 61.5 & 52.8 & 51.2 \\
& \tt Amacer\textsuperscript{-R} & 70.7 & 81.6 & 86.3 & 40.8 & 53.9 && 52.3 & 66.8 & 74.8 & 64.4 & 64.5 \\
& \tt Amacer & \bf 73.4 & \bf 83.5 & \bf 88.1 & \bf 41.3 & \bf 54.8 && \bf 55.8 & \bf 69.9 & \bf 77.3 & \bf 64.5 & \bf 66.0 \\
\midrule
\multirow{3}{*}{\shortstack[c]{\it New}} & \tt OA-Mine\textsuperscript{*} & 13.9 & 16.3 & 52.5 & 17.7 & 21.6 && 11.4 & 12.7 & 46.9 & 25.6 & 24.6 \\
& \tt Amacer\textsuperscript{-R} & 15.7 & 20.2 & 63.3 & \bf 29.1 & 31.0 && 15.6 & 19.2 & 61.6 & \bf 48.8 & 38.8 \\
& \tt Amacer & \bf 37.9 & \bf 49.1 & \bf 77.5 & 27.0 & \bf 36.2 && \bf 40.2 & \bf 52.9 & \bf 75.3 & 39.3 & \bf 46.2 \\
\bottomrule
\end{tabular}}
\caption{Decomposed evaluation results on seed attribute types only (\textit{Seed}) or on new attribute types only (\textit{New}). All models have performance degradation on new attribute types, showing that discovering open-world new attributes is a harder task than extracting seed attribute types seen in the training.}
\label{tab:result-seed-new}
% \vspace{-1.5ex}
\end{table*}

\begin{table*}[thbp!]
\centering
\resizebox{0.91\textwidth}{!}{
\begin{tabular}{ll|ccccccccccc}
\toprule
&& \multicolumn{5}{c}{Exact Match} && \multicolumn{5}{c}{Partial Match} \\
\cmidrule{3-7} \cmidrule{9-13}
&& Jaccard & ARI & NMI & Recall & \bf F1 && Jaccard & ARI & NMI & Recall & \bf F1 \\
\midrule
\multirow{3}{*}{\shortstack[c]{\it Title}} & \tt OA-Mine\textsuperscript{*} & 78.0 & 84.8 & 85.4 & 35.6 & 48.4 && 44.8 & 55.1 & 61.8 & 59.6 & 56.6 \\
& \tt Amacer\textsuperscript{-R} & 84.0 & \bf 89.1 & 88.2 & 37.6 & 50.9 && 54.1 & 64.1 & 68.4 & \bf 60.8 & 61.2 \\
& \tt Amacer & \bf 84.3 & 88.9 & \bf 90.3 & \bf 38.0 & \bf 51.5 && \bf 56.1 & \bf 65.5 & \bf 70.0 & 60.0 & \bf 61.5 \\
\midrule
\multirow{3}{*}{\shortstack[c]{\it BP}} & \tt OA-Mine\textsuperscript{*} & 52.8 & 65.0 & 71.8 & 29.6 & 39.6 && 41.9 & 53.2 & 58.6 & 47.7 & 49.0 \\
& \tt Amacer\textsuperscript{-R} & 55.2 & 66.4 & 77.2 & 35.0 & 45.3 && 46.3 & 57.7 & 66.5 & \bf 57.2 & 56.9 \\
& \tt Amacer & \bf 65.3 & \bf 75.1 & \bf 82.7 & \bf 35.2 & \bf 46.8 && \bf 54.3 & \bf 65.4 & \bf 71.9 & 56.3 & \bf 59.3 \\
\bottomrule
\end{tabular}}
\caption{Decomposed evaluation results on product titles only (\textit{Title}) or bullet point description only (\textit{BP}). All models show performance degradation on bullet point description, indicating that bullet point description has its own traits compared to titles, requiring stronger span extraction and inference.}
\label{tab:result-title-bp}
% \vspace{-1.5ex}
\end{table*}

\begin{table*}[thbp!]
\centering
\resizebox{0.55\textwidth}{!}{
\begin{tabular}{ll|ccccc}
\toprule
&& Jaccard & ARI & NMI & Recall & \bf F1 \\
\midrule
\multirow{3}{*}{\shortstack[c]{\it GOLD}} & \tt OA-Mine\textsuperscript{*} & 54.8 & 66.7 & 68.7 & 59.7 & 61.2 \\
& \tt Amacer\textsuperscript{-R} & 63.1 & 73.8 & 78.5 & \bf 68.4 & 69.8 \\
& \tt Amacer & \bf 70.0 & \bf 78.9 & \bf 83.0 & 68.2 & \bf 71.9 \\
\bottomrule
\end{tabular}}
\caption{Evaluation results by directly using gold attribute values as candidate spans. The overall evaluation of each model gets boosted as expected, and directly reflects the performance of our proposed representation learning and grouping (Section~\ref{sec:seed-attr}\&\ref{sec:new-attr}). Note that scores are the same for either exact-match or partial-match.}
\label{tab:result-gold}
% \vspace{-1.5ex}
\end{table*}

\begin{table*}[tbp!]
\centering
\resizebox{\textwidth}{!}{
\begin{tabular}{p{2cm}|lp{2cm}p{3cm}|l}
\multicolumn{2}{c}{\it \large Flavor Profile} && \multicolumn{2}{c}{\it \large Health Benefit} \\
\cmidrule{1-2} \cmidrule{4-5}
\textbf{Seed} (6) & \textbf{Extracted} && \textbf{Seed} (34) & \textbf{Extracted} \\
\toprule
\makecell[tl]{sweet \\ sweetened \\ unsweetened \\ sour \\ bitter \\ fruity} &
\makecell[tl]{nutty \\ bold \\ savory tea \\ refreshing taste \\ warm \\ great tasting beverage \\ delicious drink \\ fruit-flavored \\ floral with honey notes \\ flowery \\ earthy \\ tangy and fruity \\ tart \\ delicate flavor \\ light \\ slightly sweet and spicy \\ minty \\ hot or cold \\ savored \\ sweet and savory spice flavors \\ spicy taste \\ tasting \\ unsweetened zero calories \\ unsweetened green tea flavor \\ sweet in taste \\ crisp and refreshing \\ smooth \\ sweet and spicy taste \\ vegetal flavor \\ smokiness \\ bright and floral flavor \\ complex and rich flavors \\ aromatic \\ rich treat \\ plain \\ rich flavour \\ teas--malty \\ hearty \\ rich flavor \\ toasty texture \\ delicately floral \\ fruity flavor \\ slightly tangy \\ delicate zesty \\ accented \\ refreshingly tart herbal \\ vibrant \\ pleasantly roasted \\ bitter notes \\ ...} &&
\makecell[tl]{anti aging \\ anti-aging \\ boost energy \\ cleansing \\ cold relief \\ detox \\ detoxification \\ detoxify \\ digestive health \\ energizer \\ fertility \\ gut health \\ head relief \\ hydrated \\ immune support \\ immunity \\ laxative \\ metabolism \\ moisturize \\ mood tonic \\ nausea relief \\ night cleanse \\ nourishing \\ reduce bloating \\ relaxing herbal \\ sinus soother \\ sleep support \\ slenderizer \\ soothing \\ stress relief \\ supports immune \\ throat tamer \\ weight loss \\ weight management} &
\makecell[tl]{supports nervous system health \\ hypoallergenic \\ relieve fatigue \\ curb sugar and hunger cravings \\ thirst quenching bottle \\ help support a healthy heart \\ brighten our day \\ helps boost metabolism \\ consistency \\ lives \\ environment \\ promotes healthy liver function \\ nourishes \\ awakening \\ relieves gas and bloating \\ supports the cardiovascular system \\ supports joint health \& overall \\ experiencing the true taste \\ celebration \\ hormone balance and reproductive health \\ taste and active properties \\ helps regulate female hormone function \\ relieve menopause symptoms \\ enhance libido \\ reduce pain \\ increase fertility \\ improve mood \\ clear your head \\ yet soothing \\ helps support a healthy lifestyle \\ energy \& immunity booster \\ properties and ayurvedic benefits \\ exceptional nutritious properties \\ nutritious \\ promote healthier lifestyle choices \\ unique energy characteristics \\ reduce the jitters and crash \\ steady and prolonged alertness \\ boost cognitive function \\ body breakthrough trim \\ balanced diet plan \\ improves eye and vision health \\ five senses \\ creating health \& wellness foods \\ support healthy menstruation \\ release harmful toxins \\ morning cleanse \\ cleanse your digestive tract \\ detoxify your whole body \\ ...} \\
\end{tabular}}
\caption{Sampled predictions on TEA products of two seed attributes: \textit{Flavor Profile}, \textit{Health Benefit}. \textbf{Seed} columns display all seed values of the two attributes; \textbf{Extracted} columns show the predictions, which are extracted spans by \texttt{Amacer} from product titles or bullet point description. Given the limited amount of seed values, the model is able to expand much more diverse and long-tail expressions of attributes of interest, by up to 12 times for \textit{Flavor Profile} on the test set. Indeed, the predictions also contain certain noise, due to the lightly-supervised setting.}
\label{tab:result-example}
% \vspace{-1.5ex}
\end{table*}

% \section{Qualitative Analysis}
% \label{adpx:qual}

% Following Section~\ref{sec:qual-analysis}, detailed examples are provided from three perspectives for qualitative analysis:
% \begin{itemize}[leftmargin=*]
%     \item Table~\ref{tab:result-example-new} shows examples of extracted attribute spans on bullet point description, illustrating the model's ability of open-world attribute discovery.
%     \item Table~\ref{tab:result-example} provides sampled predictions on two seed attributes, which demonstrates strong performance on expanding diverse values for existing seed attributes.
%     \item Table~\ref{tab:result-example-topic} shows several learned latent attributes by the unsupervised topic modeling (Section~\ref{subsec:new-attr}), with top candidate spans of each attribute.
% \end{itemize}

% \paragraph{Latent Attributes} As Table~\ref{tab:result-example-topic} demonstrates that certain learned latent attributes can resemble ``concepts'', which would assist the representation learning by providing implicit semantic signals, we also observe that certain learned attributes are repetitive, such that their attribute embeddings have high cosine similarity. This behavior aligns with a previously observed issue in neural topic modeling known as \textit{topic collapsing} \cite{srivastava2017autoencoding,anonymous2023neural}, resulting in deficient topic discovery. We do not further address this issue in this work, and leave it for future research.

\end{document}